\documentclass[conference]{IEEEtran}
\IEEEoverridecommandlockouts
\usepackage{cite}
\usepackage{amsmath,amssymb,amsfonts}
\usepackage{algorithmic}
\usepackage{graphicx}
\usepackage{textcomp}
\usepackage{xcolor}
\usepackage{booktabs}
\usepackage{url}
\usepackage{multirow}
\def\BibTeX{{\rm B\kern-.05em{\sc i\kern-.025em b}\kern-.08em
    T\kern-.1667em\lower.7ex\hbox{E}\kern-.125emX}}

\begin{document}

\title{Pattern-Enhanced RT-DETR for Multi-Class Battery Detection}

\author{
\IEEEauthorblockN{Xu Zhong}
\IEEEauthorblockA{
Independent Researcher \\
xu.zhong.research@gmail.com}
\and
\IEEEauthorblockN{Enyuan Hu$^\dagger$}
\IEEEauthorblockA{
% \textit{Department Name} \\
\textit{Chemistry Division} \\
Brookhaven National Laboratory\\
NY, USA \\
enhu@bnl.gov}
\thanks{$^\dagger$Corresponding author: Enyuan Hu (enhu@bnl.gov)}
}

\maketitle

\begin{abstract}
Accurate and efficient battery detection is increasingly important
for applications in electronic waste recycling, industrial quality
control, and automated sorting systems.
In this paper, we present both a comprehensive benchmark and a
novel method for multi-class battery detection.
We systematically compare three CNN-based detectors (YOLOv8n,
YOLOv8s, YOLO11n) and two transformer-based detectors (RT-DETR-L,
RT-DETR-X) on a publicly available dataset of approximately 8,591
annotated images under identical experimental conditions, and
further propose PaQ-RT-DETR, which introduces pattern-based dynamic
query generation into RT-DETR to alleviate query activation
imbalance with negligible computational overhead.
Among baselines, YOLO11n achieves the best CNN-based accuracy
(mAP@50: 0.779) at only 2.6M parameters, while YOLOv8n delivers
the fastest inference at $\sim$1,667\,FPS.
PaQ-RT-DETR-X achieves the highest overall mAP@50 of 0.782,
surpassing RT-DETR-X by $+$2.8\% with consistent per-class gains
across all six battery categories including the data-scarce Bike
Battery class.
Our findings provide practical guidance for selecting object
detection models in battery-related industrial applications.
\end{abstract}

\begin{IEEEkeywords}
battery detection, object detection, YOLO, DETR, transformer,
deep learning, electronic waste, class imbalance
\end{IEEEkeywords}
% \vspace{-1em}

% ============================================================
\section{Introduction}
% ============================================================

\begin{figure}[t]
  \centering
  \includegraphics[width=\linewidth,height=4cm]{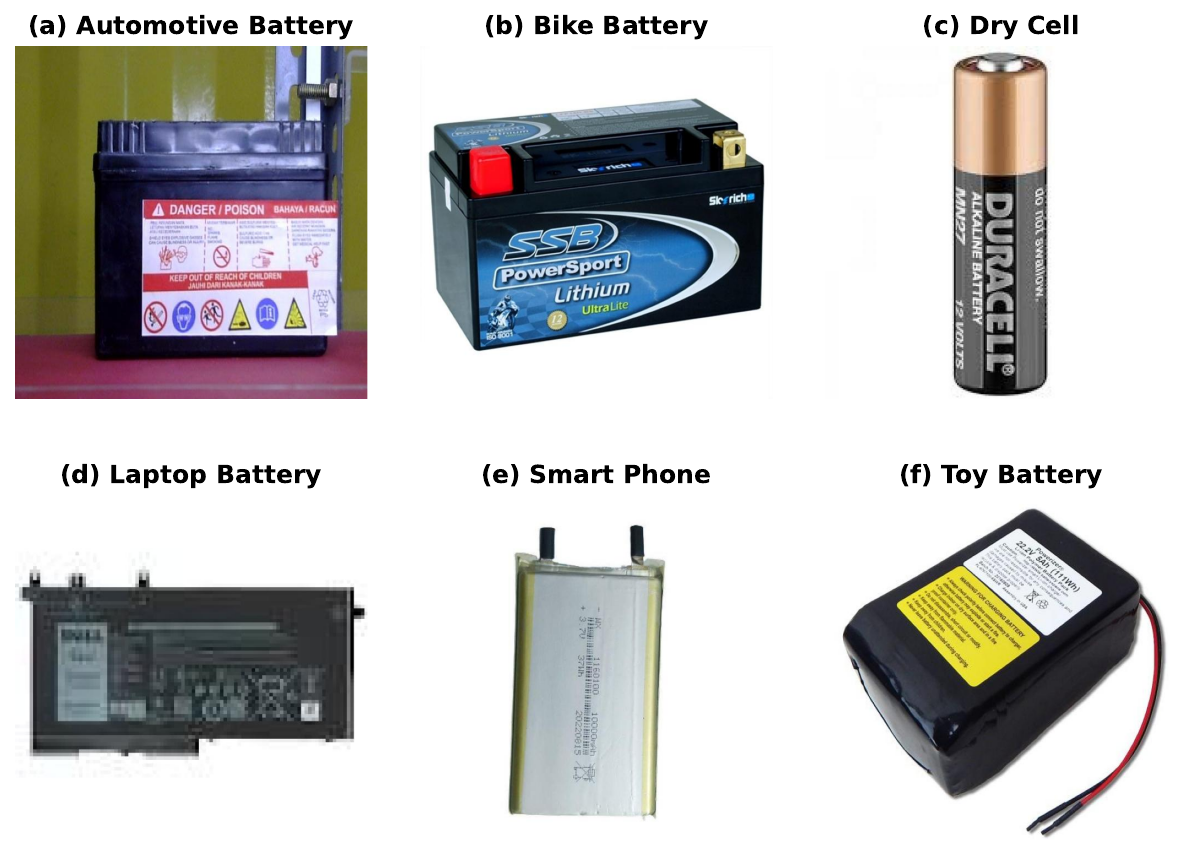}
  \caption{Representative samples from the six battery categories in our dataset.
    The dataset exhibits significant variation in battery size, shape,
    and visual appearance across categories.}
  \label{fig:battery_samples}
  \vspace{-1.5em}
\end{figure}

The rapid growth of electric vehicles and portable electronics has led to a surge in battery waste worldwide. Improper disposal poses serious environmental and safety hazards due to toxic heavy metals and flammable electrolytes \cite{b_waste}, while valuable materials such as lithium and cobalt can be recovered through proper recycling.
In modern recycling facilities, batteries arrive in mixed streams containing diverse types, each requiring different handling procedures.
As illustrated in Fig.~\ref{fig:battery_samples}, batteries exhibit diverse form factors, varying scales, and significant appearance differences across categories, posing unique challenges for automated detection.
Manual sorting is labor-intensive and exposes workers to safety risks, making automated battery detection systems essential for intelligent recycling pipelines.

Object detection has seen remarkable progress with deep learning, with two dominant paradigms emerging: CNN-based detectors exemplified by the YOLO series \cite{b_yolov8, b_yolo11}, and transformer-based end-to-end detectors built upon DETR
\cite{b_detr,b_dino,b_deformable_detr}, with RT-DETR \cite{b_rtdetr} achieving competitive real-time performance.
Despite these advances, a systematic comparison of CNN-based and transformer-based detectors on battery detection remains unexplored.
Furthermore, RT-DETR inherits a structural limitation shared by DETR variants: query activation imbalance under one-to-one Hungarian matching, where only a small fraction of queries receive meaningful gradient updates \cite{kang2026paq}.
This problem is further exacerbated by class imbalance in real-world battery datasets, where rare categories receive even fewer matching opportunities during training.

To address these challenges, we propose PaQ-RT-DETR, which introduces pattern-based dynamic query generation into RT-DETR, motivated by PaQ-DETR\cite{kang2026paq}.
By constructing decoder queries as convex combinations of shared learnable patterns, gradient updates propagate through shared parameters to benefit all queries, alleviating imbalance.
The main contributions of this paper are as follows:
\vspace{-0.5em}
\begin{itemize}
\item We present the first systematic benchmark of CNN-based (YOLOv8n, YOLOv8s, YOLO11n) and transformer-based (RT-DETR-L, RT-DETR-X) detectors on multi-class battery detection across six categories and $\sim$8,591 images.
\item We propose PaQ-RT-DETR, a lightweight enhancement of RT-DETR with pattern-based dynamic queries that alleviates query activation imbalance with negligible computational overhead.
\item We achieve 78.2\% mAP@50, with the largest gains on rare battery categories. This more balanced query optimization directly addresses a practical challenge in automated e-waste sorting for battery recycling.
\end{itemize}

% ============================================================
\section{Related Work}
\label{sec:related}
% ============================================================
\subsection{CNN-Based Object Detection}
The YOLO series~\cite{b_yolov1} represents the most prominent line of single-stage detectors, with subsequent work refining it through anchor-free designs, feature pyramid networks, and improved training strategies.
YOLOv8~\cite{b_yolov8} introduced a decoupled anchor-free head, while YOLO11~\cite{b_yolo11} further improved efficiency via a C3k2 backbone block, attaining higher mAP with $\sim$22\% fewer parameters than YOLOv8n.

\subsection{Transformer-Based Object Detection}
DETR \cite{b_detr} reformulated detection as set prediction with a transformer encoder-decoder and bipartite matching, removing hand-crafted components such as NMS, but suffered from slow convergence. Deformable DETR~\cite{b_deformable_detr} mitigated this via deformable attention, and RT-DETR~\cite{b_rtdetr} further introduced an efficient hybrid encoder, achieving real-time speeds competitive with YOLO in a fully end-to-end pipeline.
More recently, PaQ-DETR~\cite{kang2026paq} proposed pattern-based dynamic queries with quality-aware assignment to address query activation imbalance, achieving consistent mAP gains across multiple DETR backbones.

\subsection{Battery and E-Waste Detection}
Prior work on automated battery detection is limited. Ueda et al.\ \cite{b_xray} proposed an in-line sorting system combining X-ray transmission and deep learning for e-waste streams; Zhao et al.\ \cite{b_cvpr_pbd} addressed power battery quality inspection via segmentation on X-ray images; and the RecyBat24 dataset \cite{b_recybat} introduced a public benchmark for lithium-ion battery type detection. In contrast, our work focuses on RGB image-based multi-class battery detection using standard frameworks, providing a practical and reproducible baseline for real-world deployment.

% ============================================================
\section{Method}
\label{sec:method}
% ============================================================

\subsection{Preliminaries: RT-DETR}
RT-DETR formulates detection as end-to-end set prediction without NMS. An HGNetv2 backbone and a hybrid encoder produce contextualized features $Z \in \mathbb{R}^{M \times d}$ ($M$ spatial tokens, dimension $d$); the top-$K$ tokens are selected as decoder queries by classification confidence, with their spatial positions providing reference points. The transformer decoder refines these queries through $L$ layers of self- and cross-attention with $Z$:
\begin{equation}
Q^l = \text{FFN}(\text{CrossAttn}(\text{SelfAttn}(Q^{l-1}), Z))
\end{equation}
The final outputs $Q^L$ are mapped to class predictions and bounding boxes via task-specific heads.

\subsection{Overview}
\begin{figure*}[!htbp]
\centerline{\includegraphics[width=\textwidth,height=5cm]{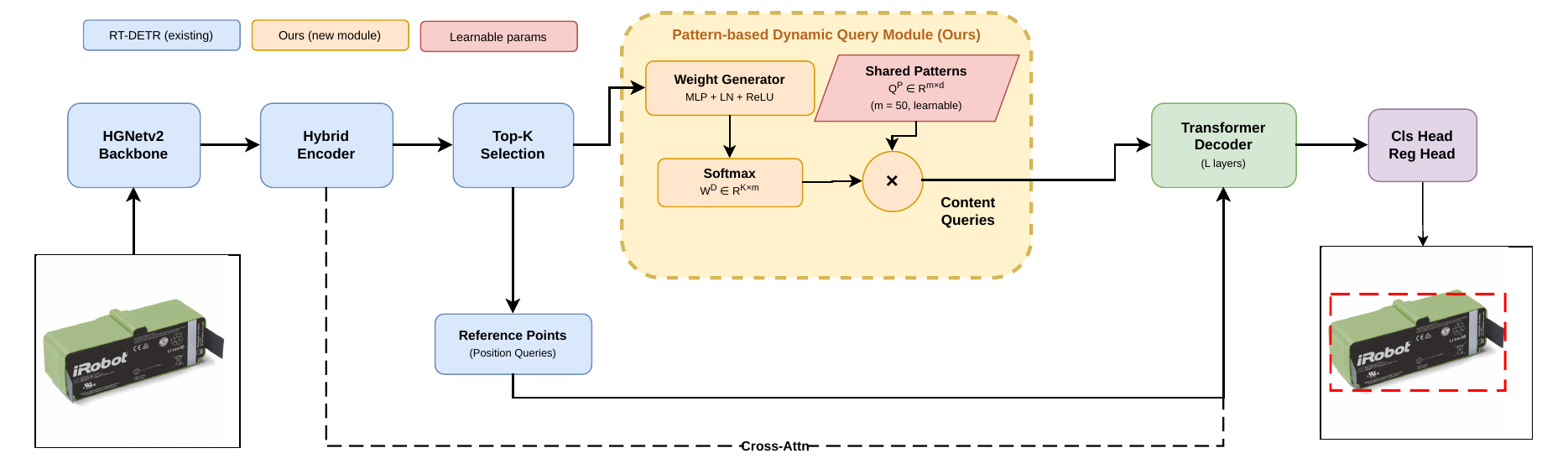}}
\vspace{-1em}
\caption{Overview of PaQ-RT-DETR. The pattern-based dynamic query module replaces RT-DETR's direct top-$K$ encoder token queries with dynamically composed content queries.}
\label{fig:architecture}
\vspace{-1.5em}
\end{figure*}

Motivated by PaQ-DETR \cite{kang2026paq}, which addresses query activation imbalance via pattern-based dynamic query generation, we propose PaQ-RT-DETR by adapting this mechanism to RT-DETR \cite{b_rtdetr}.
Here, \emph{patterns} are not hand-crafted image statistics but a small set of learnable parameter vectors in the decoder embedding space, jointly optimized end-to-end and acting as latent semantic primitives discovered from data.
Although RT-DETR's confidence-based top-$K$ encoder selection provides content-aware initialization, one-to-one Hungarian matching still causes only a small subset of queries to receive strong gradient signals---an imbalance that is especially harmful for long-tailed datasets such as battery detection. We address this by replacing the direct use of selected encoder tokens with a dynamic composition over shared learnable patterns, enabling gradient sharing across all queries; unlike PaQ-DETR \cite{kang2026paq}, the composition is integrated with RT-DETR's confidence-based top-$K$ selection. Fig.~\ref{fig:architecture} illustrates the overall architecture.

\subsection{Pattern-Based Dynamic Query Generation}
We introduce $m$ shared learnable base patterns $Q^P \in \mathbb{R}^{m \times d}$ ($m \ll K$; we set $m{=}50$, $K{=}300$). Given the top-$K$ selected encoder features $Z_\mathcal{I} \in \mathbb{R}^{K \times d}$, a lightweight two-layer MLP $F_w$ produces per-query combination weights:
\begin{equation}
W^D = \text{softmax}(F_w(Z_\mathcal{I})) \in \mathbb{R}^{K \times m}
\end{equation}
Each decoder content query is a convex combination of the shared patterns:
\begin{equation}
q_i^C = \sum_{j=1}^{m} w_{ij}^D \cdot q_j^P, \quad
w_{ij}^D \geq 0, \quad \sum_{j=1}^{m} w_{ij}^D = 1
\end{equation}
When any query is matched and updated, gradients flow back through $Q^P$, indirectly benefiting all other queries. Because all queries share the same compact basis, each pattern tends to specialize toward a recurring appearance prototype while instance-specific variation is absorbed by $W^D$, encouraging a degree of disentanglement. The composed content queries replace RT-DETR's original top-$K$ encoder features as decoder input, while position queries remain unchanged.

\subsection{Training Objective}
We follow RT-DETR's standard loss formulation
$\mathcal{L} = \mathcal{L}_{cls} + \lambda_1 \mathcal{L}_{L1} + \lambda_2 \mathcal{L}_{GIoU}$,
with default $\lambda_1, \lambda_2$, one-to-one Hungarian matching for final predictions, and auxiliary losses applied at each decoder layer.
% \vspace{-1em}

% ============================================================

\section{Experiments}
\label{sec:experiments}
% ============================================================

\subsection{Dataset}
We use the Battery Detection dataset on Roboflow Universe~\cite{b_dataset}, a publicly available large-scale dataset licensed under CC BY 4.0.
It contains approximately 8,591 RGB images annotated with bounding boxes across six battery categories: automotive battery, bike battery, dry cell, laptop battery, smartphone battery, and toy battery.
The dataset is collected from in-the-wild recycling and consumer scenes and naturally contains partial occlusion, mutual contact, viewpoint variation, complex backgrounds, mild geometric deformation (e.g., dented dry cells), and visually similar cylindrical cells (e.g., AA dry cells vs.\ 18650-style Li-ion cells in toy/laptop packs), making fine-grained discrimination non-trivial.
The dataset is pre-divided into training, validation, and test splits following an approximate 70/20/10 ratio. Table~\ref{tab:dataset} summarizes the per-class distribution in the validation split.
A notable limitation is the severe class imbalance in the Bike Battery category, which contains only 6 validation instances---far fewer than any other category.
% We retain this category to faithfully reflect real-world dataset conditions.
\vspace{-0.5cm}
\begin{table}[htbp]
\caption{Dataset Statistics (Validation Split)}
\begin{center}
\small
\begin{tabular}{llcc}
\toprule
\textbf{Class} & \textbf{Type} & \textbf{Imgs} & \textbf{Inst.} \\
\midrule
Automotive & Lead-acid      & 158 & 159 \\
Bike       & Li-ion e-bike  & 6   & 6   \\
Dry Cell   & Alkaline/Zinc  & 143 & 415 \\
Laptop     & Li-ion pack    & 52  & 60  \\
Smart Phone & Li-ion cell   & 70  & 86  \\
Toy        & Small alkaline & 185 & 226 \\
\midrule
\textbf{Total (val)} & & \textbf{692} & \textbf{952} \\
\textbf{Total (all)} & & \textbf{8,591} & \textbf{--} \\
\bottomrule
\end{tabular}
\label{tab:dataset}
\end{center}
\vspace{-4em}
\end{table}

\subsection{Baseline Models}
We evaluate five baseline models spanning two architectural paradigms. For CNN-based detectors, we include YOLOv8n and YOLOv8s~\cite{b_yolov8} with anchor-free CSPDarknet backbones, and YOLO11n~\cite{b_yolo11} which achieves higher efficiency via an improved C3k2 block.
For transformer-based detectors, we include RT-DETR-L and RT-DETR-X~\cite{b_rtdetr}, which use a hybrid encoder for multi-scale feature fusion and require no NMS post-processing. All five models use ImageNet/COCO pretrained weights and are evaluated under identical training conditions alongside our proposed PaQ-RT-DETR variants.

\subsection{Implementation Details}
All models are implemented using the Ultralytics 8.4.33 framework and trained under identical conditions to ensure a fair comparison. Training is conducted on an NVIDIA RTX 3090 (24\,GB) GPU, with PyTorch 2.8.0 and Python 3.9.25. All models are trained for 100 epochs with a batch size of 16 and input resolution of $640 \times 640$, using AdamW with a cosine learning rate schedule and ImageNet/COCO pretrained weights.
Data augmentation follows the default Ultralytics pipeline, including mosaic augmentation, random horizontal flipping, scale jitter, and HSV color space augmentation.

\subsection{Main Results}
Tables~\ref{tab:performance} and~\ref{tab:efficiency} summarize the detection performance and model efficiency of all evaluated models after 100 epochs of training.
\vspace{-0.8cm}
\begin{table}[htbp]
\caption{Detection Performance (100 Epochs)}
\begin{center}
\footnotesize
\begin{tabular}{lcccc}
\toprule
\textbf{Model} & \textbf{mAP@50} & \textbf{mAP@50:95} &
\textbf{Pre.} & \textbf{Rec.} \\
\midrule
YOLOv8n~\cite{b_yolov8}   & 0.769 & 0.677 & 0.708 & 0.706 \\
YOLOv8s~\cite{b_yolov8}   & 0.777 & 0.684 & 0.703 & 0.771 \\
YOLO11n~\cite{b_yolo11}   & 0.779 & 0.686 & 0.674 & 0.741 \\
\midrule
RT-DETR-L~\cite{b_rtdetr} & 0.685 & 0.608 & \textbf{0.734} & 0.752 \\
RT-DETR-X~\cite{b_rtdetr} & 0.754 & 0.677 & 0.715 & 0.776 \\
PaQ-RT-DETR-L  & 0.729 & 0.655 & 0.719 & 0.761 \\
PaQ-RT-DETR-X  & \textbf{0.782} & \textbf{0.692} & 0.729 & \textbf{0.788} \\
\bottomrule
\end{tabular}
\label{tab:performance}
\end{center}
\vspace{-3em}
\end{table}

\begin{table}[htbp]
\caption{Model Complexity and Inference Speed}
\begin{center}
\footnotesize
\begin{tabular}{lccc}
\toprule
\textbf{Model} & \textbf{Params (M)} & \textbf{GFLOPs} & \textbf{FPS} \\
\midrule
YOLOv8n~\cite{b_yolov8}   & 3.0  & 8.1  & $\sim$1667 \\
YOLOv8s~\cite{b_yolov8}   & 11.1 & 28.4 & $\sim$833  \\
YOLO11n~\cite{b_yolo11}   & 2.6  & 6.5  & $\sim$1667 \\
\midrule
RT-DETR-L~\cite{b_rtdetr} & 32.0 & 110  & $\sim$151  \\
RT-DETR-X~\cite{b_rtdetr} & 67.4 & 234  & $\sim$137  \\
PaQ-RT-DETR-L             & 32.5 & 110  & $\sim$125  \\
PaQ-RT-DETR-X             & 68.0 & 235  & $\sim$118  \\
\bottomrule
\end{tabular}
\label{tab:efficiency}
\end{center}
\vspace{-1em}
\end{table}
Among CNN-based models, YOLO11n achieves the best accuracy with the fewest parameters and fastest inference. YOLOv8s offers slightly lower accuracy at a higher parameter cost, while YOLOv8n provides a strong lightweight baseline.
Among transformer-based models, RT-DETR-X outperforms RT-DETR-L by $+$6.9\% mAP@50, at the cost of $2.1\times$ more parameters and marginally lower FPS. Notably, both RT-DETR variants achieve higher precision than CNN counterparts, reflecting the advantage of end-to-end architectures in suppressing false positives. Our PaQ-RT-DETR-X achieves the highest overall mAP among all evaluated models, surpassing RT-DETR-X by $+$2.8\%, with only $\sim$0.6M additional parameters ($<$1\% of the backbone, $\sim$1\,GFLOPs). PaQ-RT-DETR-L similarly improves upon RT-DETR-L, confirming that pattern-based dynamic queries consistently benefit detection across both model scales.
On RTX 3090 the module adds only $\sim$1.2\,ms per frame; with TensorRT FP16/INT8 deployment on constrained industrial hardware (e.g., NVIDIA RTX 3060, Jetson Orin NX), PaQ-RT-DETR-X runs at $\sim$15--40\,FPS, well above the 5--15\,FPS requirement of typical conveyor-belt sorting lines, with the PaQ overhead remaining $<$2\% of total inference time.

\textbf{Impact on battery recycling.} PaQ-RT-DETR-X achieves 78.2\% mAP@50 ($+$2.8\% over RT-DETR-X), with the largest gains on underrepresented categories: Bike Battery (+2.4\% relative) and Dry Cell (+3.1\% relative). These rare battery types are often the most valuable to recover yet the hardest to detect. Our results show that pattern-based queries\cite{kang2026paq} address this gap without collecting more labeled data---a costly process for rare categories.

\begin{table}[t]
\centering
\caption{Per-class performance of DETR-based detectors on the battery detection dataset (100 epochs).}
\label{tab:detr_per_class}
\setlength{\tabcolsep}{5pt}
\renewcommand{\arraystretch}{1.0}
\footnotesize
\begin{tabular}{l|cc|cc}
\toprule
\multirow{2}{*}{\textbf{Class}}
& \multicolumn{2}{c|}{\textbf{RT-DETR-X}}
& \multicolumn{2}{c}{\textbf{PaQ-RT-DETR-X}} \\
\cmidrule(lr){2-3} \cmidrule(lr){4-5}
& mAP$_{50}$ & mAP$_{50:95}$
& mAP$_{50}$ & mAP$_{50:95}$ \\
\midrule
Automotive Battery & 0.895 & 0.842 & 0.902 & 0.848 \\
Bike Battery       & 0.581 & 0.581 & 0.595 & 0.591 \\
Dry Cell           & 0.739 & 0.548 & 0.762 & 0.573 \\
Laptop Battery     & 0.871 & 0.861 & 0.890 & 0.874 \\
Smart Phone        & 0.796 & 0.758 & 0.811 & 0.773 \\
Toy Battery        & 0.641 & 0.472 & 0.656 & 0.488 \\
\midrule
\textbf{All}       & 0.754 & 0.677 & 0.782 & 0.692 \\
\bottomrule
\end{tabular}
\vspace{-1.5em}
\end{table}

\vspace{-0.3em}
\subsection{Ablation and Per-Class Analysis}
Removing the pattern module reduces PaQ-RT-DETR to vanilla RT-DETR, so Tables~\ref{tab:performance} and \ref{tab:detr_per_class} directly serve as ablation. The module yields consistent gains on both backbones ($+$4.4\% and $+$2.8\% mAP@50 on RT-DETR-L and -X respectively), with the larger gain on the smaller backbone suggesting it compensates for tighter decoder capacity. Since the cost is negligible (0.5--0.6\,M parameters, $\sim$1\,GFLOPs), the improvement cannot be attributed to added capacity.

Per-class results show that PaQ improves \emph{every} category, with the largest gains concentrated on data-scarce or visually confusable classes: Dry Cell ($+$2.3\%), Laptop ($+$1.9\%), Smart Phone ($+$1.5\%), Bike Battery ($+$1.4\%). The latter three involve cylindrical-cell families (AA dry cells vs.\ 18650-style Li-ion cells, Sec.~IV.A) where discrimination depends on subtle texture and end-cap geometry, suggesting that distinct latent patterns specialize toward such appearance variants rather than collapsing them into a single dominant cluster.

\begin{figure}[t]
\centerline{\includegraphics[width=\columnwidth,height=3cm]{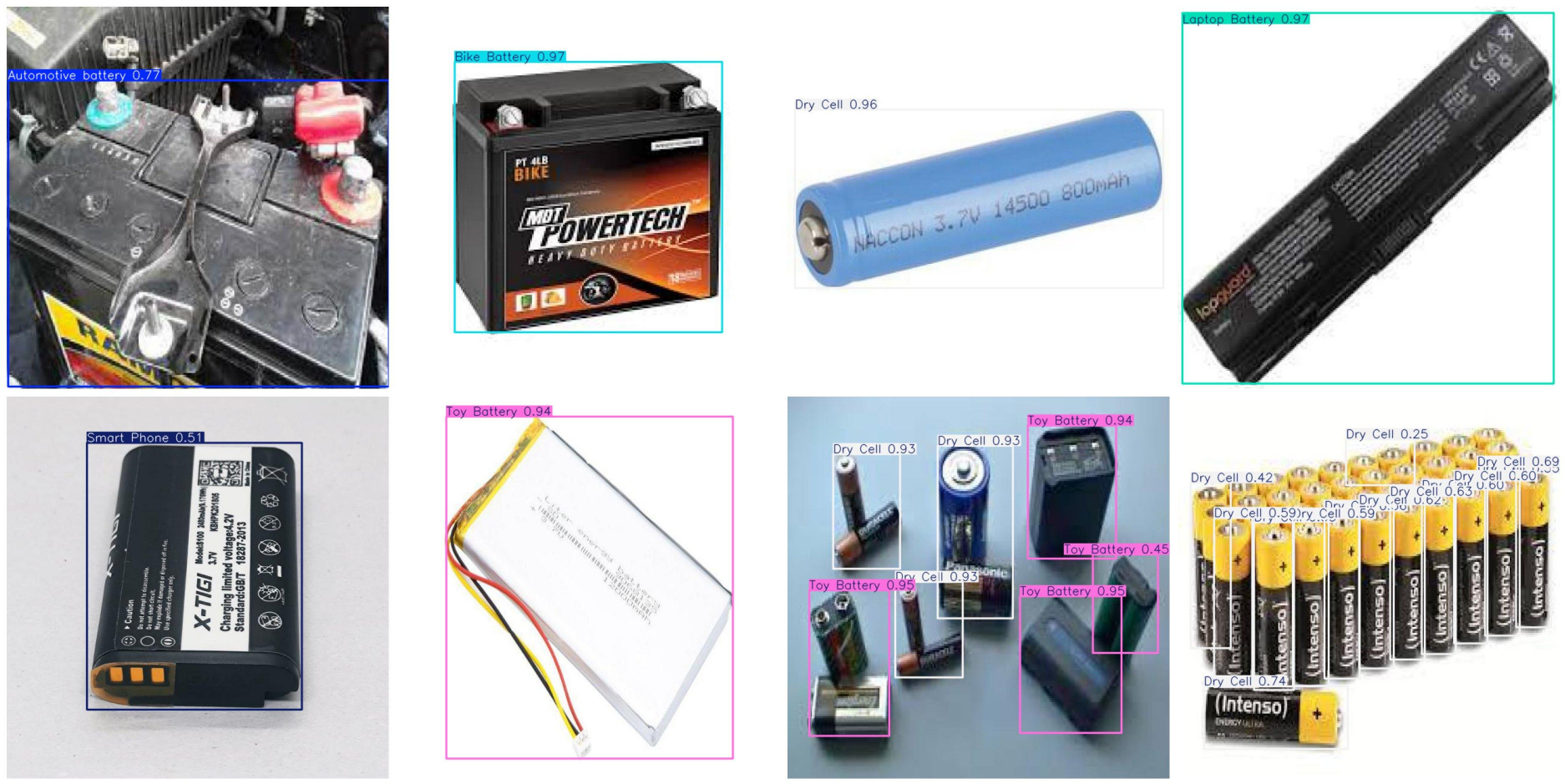}}
\caption{Qualitative results of PaQ-RT-DETR-L on the validation set,
covering cluttered multi-object scenes
including partial occlusion and mutual contact.}
\label{fig:qualitative}
\vspace{-1.5em}
\end{figure}
\subsection{Qualitative Results}
Fig.~\ref{fig:qualitative} shows detection results of PaQ-RT-DETR-L on validation
images across all six categories. The model localizes batteries robustly under
varying scales, viewpoints, and background clutter, and correctly separates
adjacent instances in multi-object scenes, including cases with partial occlusion and mild deformation, where the convex-combination structure limits the impact of locally corrupted encoder tokens. Consistent with
Table~\ref{tab:detr_per_class}, Automotive and Laptop Battery achieve the
highest confidence, while Bike Battery remains the hardest category due to
limited training samples.

% ============================================================
% \vspace{-0.5cm}
\section{Conclusion}
\label{sec:conclusion}
% ============================================================

We presented a systematic benchmark and a novel method for multi-class battery detection across six categories on $\sim$8,591 annotated images, and proposed PaQ-RT-DETR, which alleviates query activation imbalance in RT-DETR through pattern-based dynamic query composition. YOLO-based models offer strong speed-accuracy trade-offs for resource-constrained deployment, while PaQ-RT-DETR consistently improves upon RT-DETR with negligible overhead and the largest gains on data-scarce or visually confusable categories---a key requirement for recovering valuable yet rare battery types in automated recycling. In future work, we plan to investigate data augmentation strategies, additional DETR variants, and cross-modal X-ray imaging.

\section{Acknowledgement}
E.H. was supported by the Assistant Secretary for Energy Efficiency and Renewable Energy (EERE), Vehicle Technology Office (VTO) of the US Department of Energy (DOE) through the Advanced Battery Materials Research (BMR) Program under contract no. DE-SC0012704.

% ============================================================
\bibliographystyle{IEEEtran}
\bibliography{references}

\end{document}